\documentclass[letterpaper]{article} % DO NOT CHANGE THIS
\usepackage{aaai2026}
\usepackage{times}  % DO NOT CHANGE THIS
\usepackage{helvet}  % DO NOT CHANGE THIS
\usepackage{courier}  % DO NOT CHANGE THIS
\usepackage[hyphens]{url}  % DO NOT CHANGE THIS
\usepackage{graphicx} % DO NOT CHANGE THIS
\usepackage{natbib}  % DO NOT CHANGE THIS AND DO NOT ADD ANY OPTIONS TO IT
\usepackage{caption} % DO NOT CHANGE THIS AND DO NOT ADD ANY OPTIONS TO IT
\usepackage{algorithm}
\usepackage{algorithmic}
\usepackage[T1]{fontenc}

\usepackage{newfloat}
\usepackage{listings}
\DeclareCaptionStyle{ruled}{labelfont=normalfont,labelsep=colon,strut=off} % DO NOT CHANGE THIS
\floatstyle{ruled}
\newfloat{listing}{tb}{lst}{}
\floatname{listing}{Listing}
\title{IntelliProof: An Argumentation Network-based Conversational Helper for Organized Reflection}
\author{
    Kaveh Eskandari Miandoab\textsuperscript{1}\equalcontrib,
    Katharine Kowalyshyn\textsuperscript{1}\equalcontrib,
    Kabir Pamnani\textsuperscript{2}\equalcontrib,
    Anesu Gavhera\textsuperscript{1}\equalcontrib,
    Vasanth Sarathy\textsuperscript{1},
    Matthias Scheutz\textsuperscript{1}
}
\affiliations{
    \textsuperscript{\rm 1}Tufts University \\
    \textsuperscript{\rm 2}UST
        kaveh.eskandari\_miandoab, katharine.kowalyshyn, kabir.pamnani, anesu.gavhera, vasanth.sarathy, matthias.scheutz\{@tufts.edu\}

}

\begin{document}

\maketitle

\begin{abstract}
We present IntelliProof, an interactive system for analyzing argumentative essays through LLMs. IntelliProof structures an essay as an argumentation graph, where claims are represented as nodes, supporting evidence is attached as node properties, and edges encode supporting or attacking relations. Unlike existing automated essay scoring systems, IntelliProof emphasizes the user experience: each relation is initially classified and scored by an LLM, then visualized for enhanced understanding. The system provides justifications for classifications and produces quantitative measures for essay coherence. It enables rapid exploration of argumentative quality while retaining human oversight. In addition, IntelliProof provides a set of tools for a better understanding of an argumentative essay and its corresponding graph in natural language, bridging the gap between the structural semantics of argumentative essays and the user's understanding of a given text. 

%A live demo and the system are available here to try: \textbf{https://intelliproof.vercel.app}
\end{abstract}

\begin{links}
    \link{Code}{github.com/collective-intelligence-lab/intelliproof}
    \link{Demo}{intelliproof.vercel.app}
    %\link{Datasets}{https://aaai.org/example/datasets}
    %\link{Extended version}{https://aaai.org/example/extended-version}
\end{links}

% We can change these sections, just wanted to get general structure in line for draft -Katie
\section{Introduction \& Related Work}

The rise of Large Language Models (LLMs) has drastically accelerated research in computational argumentation and automated writing support. Argumentative writing is uniquely challenging, requiring a balance of claims, supporting evidence, and counterarguments within a coherent, persuasive structure. Traditional analysis methods, from rule-based systems to neural encoders, frequently struggle to capture the nuanced interrelations between claims and evidence \cite{Elaraby_Litman_2022}.

% Since the rise of Large Language Models (LLMs), the rate of research in computational argumentation and automated writing support has accelerated drastically. 
% Argumentative writing poses a unique challenge. It must balance claims, supporting evidence, and counterarguments in a coherent structure that both persuades and informs. Traditional methods for analyzing such an essay range from rule-based systems to neural encoders, but often struggle to capture the nuanced interrelations between claims and evidence \cite{Elaraby_Litman_2022}. 

\begin{figure*}[h]
    \centering
    \includegraphics[width=0.71\linewidth]{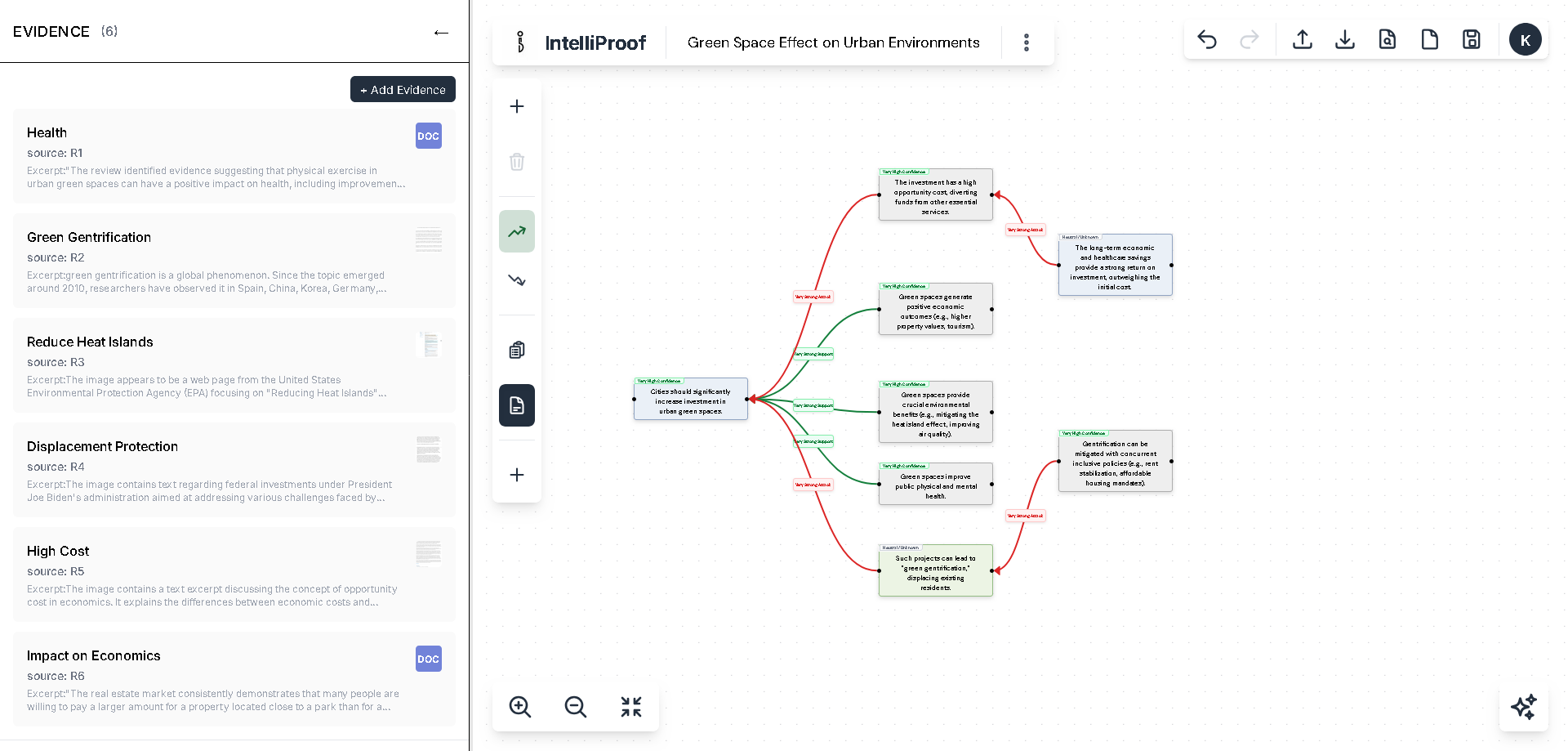}
    \caption{IntelliProof user interface overview using an example graph on the effect of green space on urban environments. }
    \label{fig:full_screen}
\end{figure*}

We introduce \textit{IntelliProof}, an LLM-powered tool that analyzes arguments by modeling them as graphs \cite{saveleva-etal-2021-graph}. In this model, claims are represented as nodes, with their strength quantified by evidence encoded as node properties. Weighted edges denote \textit{support} or \textit{attack} relations between claims. An LLM is used to score, classify, and justify these relations, while allowing human overrides for transparency and control. The dynamic identification and visualization of these relationships are shown in Figure \ref{fig:full_screen}.

By transforming essays into structured argumentation graphs, IntelliProof aims to make argumentative reasoning more interpretable, providing writers and educators with insights into essay coherence and persuasiveness. This approach contributes to the discussions on how to integrate LLMs into workflows that demand interpretability, reliability, and pedagogical value simultaneously.

%rather than sacrificing one of these priorities over another.  

%Argument mining (AM) focuses on extracting argumentative components such as claims, premises, and their relations \cite{Li_Schlegel_Sun_Batista-Navarro_Nenadic_2025a}. 
%because LLMs have been shown to outperform traditional baselines in tasks such as argument classification and counter-speech generation \cite{Chen_Cheng_Luu_Bing_2024}.

LLMs have shifted argument mining methods from encoder-based architectures to prompting and fine-tuning strategies \cite{cabessa2024argument, Favero_Perez-Ortiz_Käser_Oliver_2025}. However, annotation bottlenecks and evaluation challenges remain \cite{Schaefer_2025}. Recent work also explores interactive systems that combine generative models with human input for constructing argument graphs \cite{Lenz_Bergmann_2025}. IntelliProof extends this work by integrating graph-based structuring directly into analysis while grounding scoring of arguments in quantifiable, mathematical metrics.  

Educational applications increasingly use LLMs for essay scoring and feedback \cite{Kim_Jo_2024, Chu_Kim_Wong_Yi_2025}. Although many approaches optimize predictive accuracy, few address the interpretability of argumentative quality. Surveys of persuasive applications highlight both the promise and ethical risks of LLM-driven reasoning systems \cite{Rogiers_Noels_Buyl_Bie_2024}. By grounding essay feedback in explicit argument graphs, IntelliProof contributes to more interpretable educational tools, which will lead to safer AI systems deployed in educational settings.  

% LLMs are increasingly used for educational essay scoring and feedback \cite{Kim_Jo_2024, Chu_Kim_Wong_Yi_2025}. While accuracy is the focus, many approaches overlook argumentative interpretability. Surveys highlight the promise and ethical risks of deploying LLM-driven reasoning systems \cite{Rogiers_Noels_Buyl_Bie_2024}. By grounding feedback in explicit argument graphs, IntelliProof offers more interpretable educational tools, leading to safer AI deployment.

%The reliability of LLMs in evaluation tasks has been critically examined under the emerging ``LLM-as-a-Judge'' paradigm \cite{Gu_Jiang_Shi_Tan_Zhai_Xu_Li_Shen_Ma_Liu_etal._2025, Li_Dong_Chen_Su_Zhou_Ai_Ye_Liu_2024}. While such models excel at scalable assessments, concerns about consistency, bias, and generalizability persist \cite{Cao_Hong_Li_Ying_Ma_Liang_Liu_Yao_Wang_Huanget_al._2025, Gao_Hu_Yin_Ruan_Pu_Wan_2025}. IntelliProof mitigates these concerns by using an ensemble method of an LLM-as-a-judge and a semantic negentropy score weighed by a confidence score \textbf{update this if we need to change it based on your implementation}, joining the argument for a combination of efficiency via LLM-based evaluation along with quantifiable scores to ensure grounded, accurate results. The following sections outline our implementation of our system, which, to the best of our knowledge, is the first system that combines graph-based argumentation graphs via an LLM with modifiable features. Our demo can be found at \textbf{DEMO LINK}.

\section{Intelliproof Overview}
Intelliproof's functionality spans argument creation, scoring, classification, and generation techniques. %Each of the features elaborated on below is integrated within our GUI front-end. 

%\subsection{Argumentation Graphs}
\paragraph{Graph Visualization}
IntelliProof is designed to structurally visualize argumentative essays while providing an LLM-powered (GPT-4o for the instance of the demo given its performance \cite{shahriar2024puttinggpt4oswordcomprehensive}) toolset for the analysis of the claims. As such, users can input claims, classify them (into Fact, Policy, or Value), and establish connections between the claims via the main GUI of the tool.  

\paragraph{LLM Document Analysis}
To establish claims, users upload supporting documents as evidence in PDF or image format. A dedicated LLM instance then processes these files, suggesting relevant text or image extracts for a specific claim. The user attaches the suggested evidence to the claim via a drag-and-drop interface, which in turn prompts the LLM to assess the claim's strength by analyzing all attached evidence. Any number of supporting or negating evidence pieces can be associated with a single claim.

\paragraph{Claim Credibility Score}
To assess overall claim strength, we combine evidence and edge scores to obtain the claim credibility score $S_t$ where $S_t = \tanh(\Delta \space \frac{1}{n}\Sigma_{i=0}^{n} f_E(e_i) + \Sigma_{j=0}^m f_{ED}(k_j) * S_{t-1})$. $f_E$ and $f_{ED}$ are calculated based on the LLMs assessment of claim support based on an evidence, and based on an incoming edge, respectively, and $\Delta$ is a tunable hyperparameter. Note that given the weakness of LLMs in directly generating scores \cite{schroeder2025trustllmjudgmentsreliability, cui2025llmsscorersrethinkingmt}, we first generate a qualitative classification as the LLM's assessment, and then utilize the Evans coefficient interpretation \cite{Evans1996} to convert the qualitative assessment to numerical scores. 

\paragraph{Report Generation}
Another feature of Intelliproof is automatic report generation from the graph implementation of an argument. These reports combine evidence evaluation, edge validation, assumptions analysis,
and graph critique into a singular unified report. Our system processes graph structure,
evidence quality, relationship strengths, and logical
patterns simultaneously and creates an eight section, comprehensive report of the argumentative essay. 

%\subsection{Additional Functionalities}
\paragraph{AI Copilot Chat Interface}
Using our integrated chatbot, natural language queries are parsed, and one may ask questions
about the graph. The responding AI
is context aware, and users can get insights on
arguments' strengths, weaknesses, and gaps to fill. 
As arguments are built, the LLM context window is also updated in real-time to contain the new information.

\paragraph{Assumption Generation}

Intelliproof analyzes claim relationships to identify three implicit assumptions that would strengthen support between claims. It also finds hidden premises and bridges assumptions needed to make arguments more robust. Each assumption includes an importance rating and a justification for why it strengthens the relationship generated based on a few-shot learning approach \cite{brown2020languagemodelsfewshotlearners} prepared by an argumentation field expert. 

\paragraph{Critique Graph} 
% We additionally strive to uncover the essay's weaknesses by deploying a state-of-the-art LLM (GPT-4o), which matches the overall argument against a comprehensive bank of argument patterns and logical fallacies. Our Argument Patterns Bank is a built-in YAML database containing fallacious patterns, good arguments, and absurd reasoning examples developed by an argumentation field expert.  In this way, we are able to identify circular reasoning, straw man arguments, false causes, and other logical fallacies.

To identify essay weaknesses, we deploy a state-of-the-art LLM (GPT-4o) to match the overall argument against our comprehensive Argument Patterns Bank. This Bank is a built-in YAML database, developed by an argumentation expert, containing patterns for logical fallacies, good arguments, and absurd reasoning. This process allows us to specifically identify issues like circular reasoning, straw man arguments, and false causes.

\section{System Implementation}

IntelliProof's architecture consists of three core components. The \textbf{frontend} is built with \textit{Vite} and \textit{React.js} to create a dynamic user interface that handles all back-end API and database calls. The \textbf{backend} uses a \textit{Python} server with \textit{FastAPI} for handling requests and \textit{SupaBase} (PostgreSQL) for managing user data such as profiles, evidence files, and graphs. For the large language model, we utilize GPT-4o via OpenAI's Python library, chosen for its balance of performance, cost, and availability. The design is modular, allowing GPT-4o to be easily substituted with other locally or remotely deployed LLMs.

%\footnote{We publicly release the source code for IntelliProof at \textbf{https://github.com/collective-intelligence-lab/intelliproof}}

%outline API calls

\section{Conclusion}

Intelliproof is an interactive LLM platform that creates argument graphs based on provided evidence. This system is designed to be helpful in devising strong arguments, filling gaps in arguments, and utilizing an LLM to provide a detail-oriented look at an argumentative essay. While this system may be expanded further in the future, at present, we provide a robust, functional system that demonstrates the feasibility of Intelliproof as a powerful tool for structured, LLM-driven argumentation.

\section{Acknowledgments}

This research was supported in part by Other Transaction award HR00112490378 from the U.S. Defense Advanced Research Projects Agency (DARPA) Friction for Accountability in Conversational Transactions (FACT) program.

\bibliography{aaai2026}

\end{document}